\documentclass[runningheads]{llncs}
\usepackage{graphicx}
\usepackage{subfigure}
\usepackage{amssymb}
\usepackage{amsmath}
\usepackage{booktabs}
\usepackage{array}
\usepackage{float}
\usepackage{cite}
\usepackage[colorlinks,linkcolor=blue,citecolor=blue,urlcolor=blue]{hyperref}
\begin{document}
\title{Self-Supervised Correction Learning for Semi-Supervised Biomedical Image Segmentation}
\titlerunning{Self-Supervised Correction Learning}
\author{ Ruifei Zhang \inst{1}  \and %\textsuperscript{(\Letter)} \orcidID{0000-1111-2222-3333} 
Sishuo Liu\inst{2} \and
Yizhou Yu\inst{2,3} \and
Guanbin Li\inst{1,4}\thanks{Corresponding author is Guanbin Li~(liguanbin@mail.sysu.edu.cn).}
}

% index{Zhang, Ruifei}
% index{Liu, Sishuo}
% index{Yu, Yizhou}
% index{Li, Guanbin}

%
\authorrunning{Zhang et al.}
%\institute{No Institute Given}
\institute{
 $^1$%School of Computer Science and Engineering, 
 Sun Yat-sen University, Guangzhou, China\\ 
 $^2$The University of Hong Kong, Pokfulam, Hong Kong\\
 $^3$Deepwise AI Lab, Beijing, China\\
 $^4$Shenzhen Research Institute of Big Data, Shenzhen, China}

%\author{First Author\inst{1}\orcidID{0000-1111-2222-3333} \and
%Second Author\inst{2,3}\orcidID{1111-2222-3333-4444} \and
%Third Author\inst{3}\orcidID{2222--3333-4444-5555}}
%\institute{Princeton University, Princeton NJ 08544, USA \and
%Springer Heidelberg, Tiergartenstr. 17, 69121 Heidelberg, Germany
\maketitle              % typeset the header of the contribution
\begin{abstract}
Biomedical image segmentation plays a significant role in computer-aided diagnosis. However, existing CNN based methods rely heavily on massive manual annotations, which are very expensive and require huge human resources. In this work, we adopt a coarse-to-fine strategy and propose a self-supervised correction learning paradigm for semi-supervised biomedical image segmentation. Specifically, we design a dual-task network, including a shared encoder and two independent decoders for segmentation and lesion region inpainting, respectively. In the first phase, only the segmentation branch is used to obtain a relatively rough segmentation result. In the second step, we mask the detected lesion regions on the original image based on the initial segmentation map, and send it together with the original image into the network again to simultaneously perform inpainting and segmentation separately. For labeled data, this process is supervised by the segmentation annotations, and for unlabeled data, it is guided by the inpainting loss of masked lesion regions. Since the two tasks rely on similar feature information, the unlabeled data effectively enhances the representation of the network to the lesion regions and further improves the segmentation performance. Moreover, a gated feature fusion (GFF) module is designed to incorporate the complementary features from the two tasks. Experiments on three medical image segmentation datasets for different tasks including polyp, skin lesion and fundus optic disc segmentation well demonstrate the outstanding performance of our method compared with other semi-supervised approaches. The code is available at \url{https://github.com/ReaFly/SemiMedSeg}.

%\keywords{First keyword  \and Second keyword \and Another keyword.}
\end{abstract}
\section{Introduction}
Medical image segmentation is an essential step in computer-aided diagnosis. In practice, clinicians use various types of images to locate lesions and analyze diseases. An automated and accurate medical image segmentation technique is bound to greatly reduce the workload of clinicians.

With the vigorous development of deep learning, the FCN~\cite{long2015fully}, UNet~\cite{ronneberger2015u} and their variants~\cite{li2016deep,zhang2020adaptive} have achieved superior segmentation performance for both natural images and medical images. However, these methods rely heavily on labeled data, which is time-consuming to acquire especially for medical images. Therefore, many studies adopt semi-supervised learning to alleviate this issue, including GAN-based methods~\cite{hung2018adversarial,zhang2017deep}, consistency training~\cite{ouali2020semi,tarvainen2017mean}, pseudo labeling~\cite{lee2013pseudo} and so on. For instance, Mean Teacher (MT)~\cite{tarvainen2017mean} and its variants~\cite{yu2019uncertainty,li2020transformation} employ the consistency training for labeled data and unlabeled data by updating teacher weights via an exponential moving average of consecutive student models. Recently, some works~\cite{li2020self,chaitanya2020contrastive} integrate self-supervised learning such as jigsaw puzzles~\cite{noroozi2016unsupervised} or contrastive learning~\cite{chen2020simple} to semi-supervised segmentation and achieve competitive results. However, few of them try to dig deeply into the context and structural information of unlabeled images to supplement the semantic segmentation.

In this work, we also consider introducing self-supervised learning to semi-supervised segmentation. In contrast to ~\cite{li2020self,chaitanya2020contrastive}, we make full use of massive unlabeled data to exploit image internal structure and boundary characteristics by utilizing pixel-level inpainting as an auxiliary self-supervised task, which is combined with semantic segmentation to construct a dual-task network. As the inpainting of normal non-lesion image content will only introduce additional noise for lesion segmentation, we design a coarse-to-fine pipeline and then enhance the network's representations with the help of massive unlabeled data in the correction stage by only masking the lesion area for inpainting based on the initial segmentation result. Specifically, in the first phase, only the segmentation branch is used to acquire a coarse segmentation result, while in the second step, the masked and original images are sent into the network again to simultaneously perform lesion region inpainting and segmentation separately. Since the two tasks rely on similar feature information, we also design a gated feature fusion (GFF) module to incorporate complementary features for improving each other. Compared with the most related work~\cite{chen2019multi} which introduces a reconstruction task for unlabeled data, their two tasks lack deep interaction and feature reuse, thus cannot collaborate and facilitate each other. Besides, our network not only makes full use of massive unlabeled data, but also explores more complete lesion regions for limited labeled data through the correction phase, which can be seen as “image-level erase~\cite{wei2017object}” or “reverse attention~\cite{chen2018reverse}”.

Our contribution is summarized as follows. (1) We propose a novel self-supervised semi-supervised learning paradigm for general lesion region segmentation of medical imaging, and verify that the pretext self-supervised learning task of inpainting the lesion region at the pixel level can effectively enhance the feature learning and greatly reduce the algorithm's dependence on large-scale dense annotation. (2) We propose a dual-task framework for semi-supervised medical image segmentation. Through introducing the inpainting task, we create supervision signals for unlabeled data to enhance the network's representation learning of lesion regions and also exploit additional lesion features for labeled data, thus effectively correct the initial segmentation results. (3) We evaluate our method on three tasks, including polyp, skin lesion and fundus optic disc segmentation, under a semi-supervision setting. The experimental results demonstrate that our method achieves superior performance compared with other state-of-the-art semi-supervised methods.

\section{Methodology}
\subsection{Overview}
\begin{figure}[htp]
\centering
\includegraphics[width=0.8\textwidth]{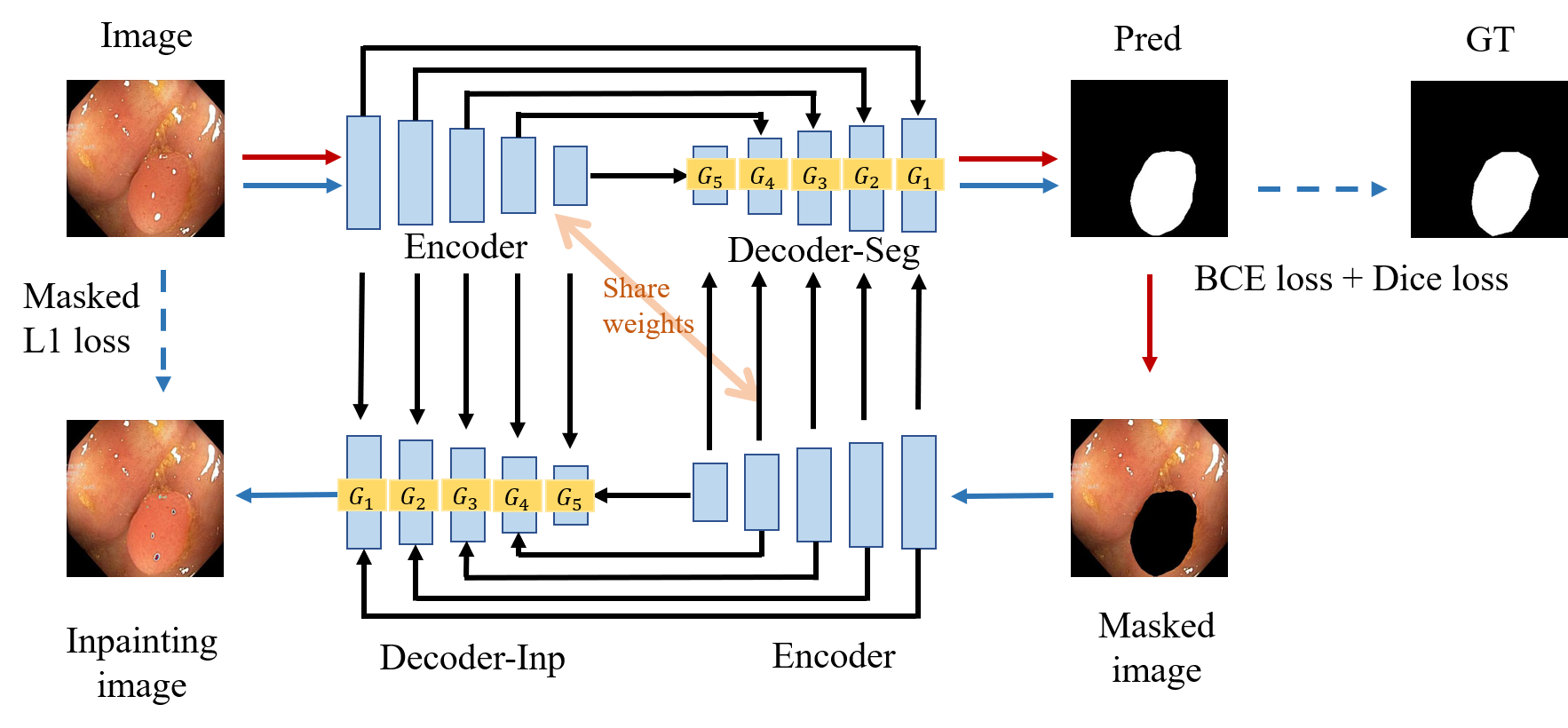} % Reduce the figure size so that it is slightly narrower than the column. Don't use precise values for figure width.This setup will avoid overfull boxes.
\caption{The overview of our network. Both encoders share weights. $G_1$-$G_5$ represent five GFF modules. The red and blue arrows denote the input and output of our network in the first and second stage respectively.}
\label{fig2}
\end{figure}

In this work, we adopt a coarse-to-fine strategy and propose a self-supervised correction learning paradigm for semi-supervised biomedical image segmentation. Specifically, we introduce inpainting as the pretext task of self-supervised learning to take advantage of massive unlabeled data and thus construct a dual-task network, as shown in Fig.~\ref{fig2}. Our proposed framework is composed of a shared encoder, two decoders and five GFF modules placed on each layer of both decoders. We utilize ResNet34~\cite{he2016deep} pretrained on the ImageNet~\cite{deng2009imagenet} as our encoder, which consists of  five blocks in total. Accordingly, the decoder branch also has five blocks. Each decoder block is composed of two Conv-BN-ReLU combinations. For the convenience of expression, we use ${E}_{seg}$, ${D}_{seg}$ to represent the encoder and decoder of the segmentation branch, and ${E}_{inp}$ and ${D}_{inp}$ for those of the inpainting branch.

In the first step, given the image $x\in\mathbb{R}^{H\times W \times C}$, in which $H$,$W$,$C$ are the height, width and channels of the image respectively, we use the segmentation branch ${E}_{seg}$, ${D}_{seg}$ with skip-connections, the traditional U-shape encoder-decoder structure, to obtain a coarse segmentation map $\hat{y}_{coarse}$ and then mask the original input based on its binary result $\overline{y}_{coarse}$ by the following formulas:

\begin{equation}
\hat{y}_{coarse} = D_{seg}(E_{seg}(x))
\label{equ1}
\end{equation}

\begin{equation}
{x}_{mask}= x \times (1-\overline{y}_{coarse})
\label{equ2}
\end{equation} 

In the second phase, the original image $x$ and the masked image ${x}_{mask}$ are sent into ${E}_{seg}$ and ${E}_{inp}$ simultaneously to extract features ${e}_{seg}$ and ${e}_{inp}$. Obviously, ${e}_{seg}$ is essential for the inpainting task, and since the initial segmentation is usually inaccurate and incomplete,  ${e}_{inp}$ may also contain important residual lesion features for the correction of the initial segmentation. In order to adaptively select the useful features of  ${e}_{inp}$ and achieve complementary fusion of ${e}_{seg}$ and  ${e}_{inp}$, we design the GFF modules ($G_1$-$G_5$) and place them on each decoder layer of both branches. Specifically, for the $i^{th}$ layer, the features ${e}_{seg}^i$ and ${e}_{inp}^i$ are delivered into ${G}_i$ through skip-connections to obtain the fusion ${e}^i$ = $G_i$(${e}_{seg}^{i}$, ${e}_{inp}^{i}$), and then sent to the corresponding decoder layer. Thus, both $G_i$ of the two branches shown in Fig.~\ref{fig2} actually share parameters, taking the same input and generating the identical output. To enhance the learning of the GFF modules, we adopt a deep supervision strategy and each layer of the two decoder branches generate a segmentation result and an inpainting result respectively by the following formulas:

\begin{equation}
\hat{y}_{fine}^i = \left\{
\begin{aligned}
D_{seg}^i([{e}^i, d_{seg}^{i+1}])&, \quad i=1,2,3,4\\
D_{seg}^i({e}^i)&, \quad i=5
\end{aligned}
\right.
\label{equ3}
\end{equation}

\begin{equation}
\hat{x}^i = \left\{
\begin{aligned}
D_{inp}^i([{e}^i, d_{inp}^{i+1}])&,\quad i=1,2,3,4\\
D_{inp}^i({e}^i)&,\quad i=5
\end{aligned}
\right.
\label{equ4}
\end{equation}

Where [$\cdot$ , $\cdot$] denotes the concatenation process, and $d_{seg}^{i+1}$, $d_{inp}^{i+1}$ represent the features from previous decoder layers. The deep supervision strategy can also avoid ${D}_{inp}$ directly copying the features of the low-level ${e}_{seg}$ to complete the inpainting task without in-depth lesion feature mining. The output of the last layer $\hat{y}_{fine}^{1}$ is the final segmentation result of our network.

\begin{figure}[htp]
\centering
\includegraphics[width=0.6\textwidth]{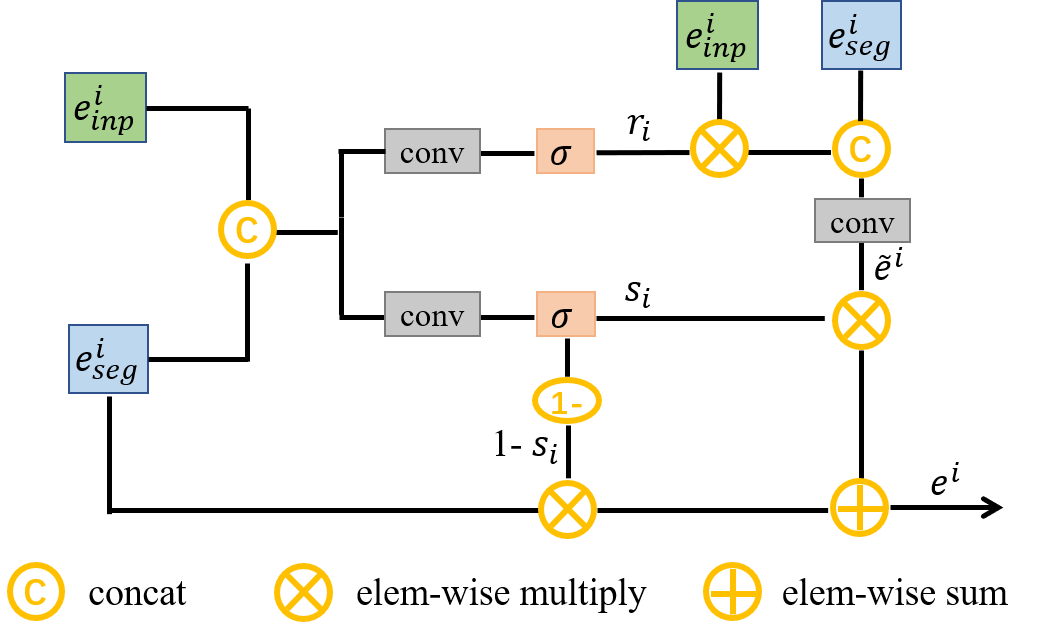} % Reduce the figure size so that it is slightly narrower than the column. Don't use precise values for figure width.This setup will avoid overfull boxes.
\caption{Gated Feature Fusion Module}
\label{fig3}
\end{figure}

\subsection{Gated Feature Fusion (GFF)}
To better incorporate complementary features and filter out the redundant information, we design the GFF modules placed on each decoder layer to integrate the features delivered from the corresponding encoder layer of two branches. The details are shown in Fig.~\ref{fig3}. Our GFF module consists of a reset gate and a select gate. Specifically, for the $G_{i}$ placed on the $i^{th}$ decoder layer, the value of two gates is calculated as follows:
\begin{equation}
r_{i}= \sigma(W_{r}\left [{e}_{seg}^{i}, {e}_{inp}^{i} ] \right )
\label{equ5}
\end{equation} 
\begin{equation}
s_{i}= \sigma(W_{s}\left [{e}_{seg}^{i}, {e}_{inp}^{i} ] \right )
\label{equ6}
\end{equation} 
Where $W_{r}$, $W_{s}$ denote the convolution process, taking the concatenation of ${e}_{seg}^{i}$ and ${e}_{inp}^{i}$ as input. $\sigma$ represents the Sigmoid function. ${r}_{i}$ and ${s}_{i}$ represent the value of the reset gate and the select gate, respectively. Since the input of the inpainting branch is the masked image, the reset gate is necessary to suppress massive invalid background information. And then the select gate achieves adaptive and complementary feature fusion between the reintegrated features $\tilde{e}^i$ and the original segmentation feature ${e}_{seg}^{i}$ by the following operations:
\begin{equation}
\tilde{e}^i = W\left [{r}_{i}\times {e}_{inp}^{i}, {e}_{seg}^{i} \right ]  )
\label{equ7}
\end{equation} 
\begin{equation}
{e}^i = s_{i} \times \tilde{e}^i + (1-s_{i}) \times {e}_{seg}^{i}
\label{equ8}
\end{equation} 
where $W$ also represents the convolution process to make the reintegrated features $\tilde{e}^i$ have the same dimension with ${e}_{seg}^{i}$.

\subsection{Loss Function}
We only calculate loss in the second stage. The labeled dataset and unlabeled dataset are denoted as $D_l$ and $D_u$. For the labeled data $x_l \in D_l$, $y_l$ is the Ground Truth. Since we adopt the deep supervision strategy, the overall loss is the sum of the combination of Binary CrossEntropy (BCE) loss and Dice loss between each output and the Ground Truth:
\begin{equation}
\mathcal{L}_{seg}(x_l) = \sum\limits_{i=1}^5{\mathcal{L}_{BCE}^i(\hat{y}_{l}^i, y_l^i) + \mathcal{L}_{Dice}^i(\hat{y}_{l}^i,y_l^i)}
\label{equ9}
\end{equation} 
where $\hat{y}_{l}^i$, $y_l^{i}$ denote the segmentation map $\hat{y}_{fine}^i$ of the $i^{th}$ decoder layer and the corresponding down-sampling Ground Truth $y_l$.

For unlabeled data $x_u \in D_u$, the inpainting loss is the sum of $L1$ loss between each inpainting image and the original image in the masked region: 
\begin{equation}
\mathcal{L}_{inp}(x_u) =  \sum\limits_{i=1}^5{\overline{y}_{u}^i\times \left| \hat{x}_u^i - x_u^i \right|}
\label{equ12}
\end{equation}
where $\hat{x}_u^i$, $x_u^i$ and $\overline{y}_{u}^i$ represent the inpainting image, down-sampling original image and binary segmentation result of the $i^{th}$ decoder layer, respectively. In the end, the total loss function is formulated as follows: 
\begin{equation}
\mathcal{L}= \lambda_1\sum\limits_{x_l\in D_l}{\mathcal{L}_{seg}(x_l)} + \lambda_2\sum\limits_{x_u \in D_u}{\mathcal{L}_{inp}(x_u)}
\label{equ13}
\end{equation}
where $\lambda_1$, $\lambda_2$ are weights balancing the segmentation loss and the inpainting loss. And we set $\lambda_1=2$ and $\lambda_2=1$ in our experiments.

\section{Experimental Results}

\subsection{Dataset and Evaluation Metric}
We conduct experiments on a variety of medical image segmentation tasks to verify the effectiveness and robustness of our approach, including polyp, skin lesion and fundus optic disc segmentation, respectively.\\
\textbf{Polyp Segmentation}
We use the publicly available kvasir-SEG~\cite{jha2020kvasir} dataset containing 1000 images, and randomly select 600 images as the training set, 200 images as the validation set, and the remaining as the test set. \\
\textbf{Skin Lesion Segmentation}
We utilize the ISBI 2016 skin lesion dataset~\cite{gutman2016skin} to evaluate our method performance. This dataset consists of 1279 images, among which 900 are used for training and the others for testing.\\
\textbf{Optic Disc Segmentation}
The Rim-one r1 dataset~\cite{fumero2011rim} is utilized in our experiments, which has 169 images in total. We randomly split the dataset into a training set and a test set with the ratio of 8:2.\\
\textbf{Evaluation Metric}
Referring to common semi-supervised segmentation settings~\cite{yu2019uncertainty,li2020transformation}, for all datasets, we randomly use $20\%$ of the training set as the labeled data, $80\%$ as the unlabeled data and adopt five metrics to quantitively evaluate the performance of our approach and other methods, including “Dice Similarity Coefficient (Dice)”, “Intersection over Union (IoU)”, “Accuracy (Acc)”, “Recall (Rec)” and “Specificity (Spe)”.

\subsection{Implementation Details}
\textbf{Data pre-processing}
In our experiments, since the image resolution of all datasets varies greatly, we uniformly resize all images into a fixed size of $320\times320$ for training and testing. And in the training stage, we use data augmentation, including random horizontal and vertical flips, rotation, zoom, and finally all the images are randomly cropped to $256\times256$ as input.\\
\textbf{Training details}
Our method is implemented using PyTorch~\cite{paszke2019pytorch} framework. We set batch size of the training process to 4, and use SGD optimizer with a momentum of 0.9 and a weight decay of 0.00001 to optimize the model. A poly learning rate policy is adopted to adjust the initial learning rate, which is $lr=init\_lr\times(1-\frac{iter}{max\_iter})^{power}$, where $init\_lr=0.001$, $power=0.9$). The total number of epochs is set to 80.

\begin{table}
\centering
\caption{Comparison with other state-of-the-art methods and ablation study on the Kvasir-SEG dataset}
\resizebox{\textwidth}{!}{
\begin{tabular}{p{80pt}|p{80pt}|p{30pt}|p{30pt}|p{30pt}|p{30pt}|p{30pt}} % 控制表格的格式
\toprule
Methods & Data & $Dice$ & $IoU$ & $Acc$ & $Rec$ & $Spe$ \\
\midrule
Supervised &600L (All) &89.48&83.69&97.34&91.06&98.58 \\
Supervised &120L &84.40&76.18&96.09&85.35&98.55 \\
\hline
DAN~\cite{zhang2017deep} &120L + 480U&85.77&78.12&96.37&86.86&98.53 \\
MT~\cite{tarvainen2017mean} &120L + 480U &85.99&78.84&96.21&86.81&\textbf{98.79} \\
UA-MT~\cite{yu2019uncertainty}&120L + 480U &85.70&78.34&96.38&88.51&98.40 \\
TCSM\_V2~\cite{li2020transformation}&120L + 480U &86.17&79.15&96.38&87.14&98.76\\
MASSL~\cite{chen2019multi}&120L + 480U&86.45&79.61&96.34&89.18&98.32 \\
\textbf{Ours} &120L + 480U &\textbf{87.14}&\textbf{80.49}&\textbf{96.42}&\textbf{90.78}&97.89 \\
\hline
Ours~(add)&120L + 480U&85.59&78.56&96.12&87.98&98.26\\
Ours~(concat)&120L + 480U&86.09&78.98&96.21&90.54&97.63\\
\bottomrule
\end{tabular}}
\label{tab1}
\end{table}

\subsection{Comparisons with the State-of-the-Art}
\begin{figure}[htp]
\centering
\includegraphics[width=0.9\textwidth]{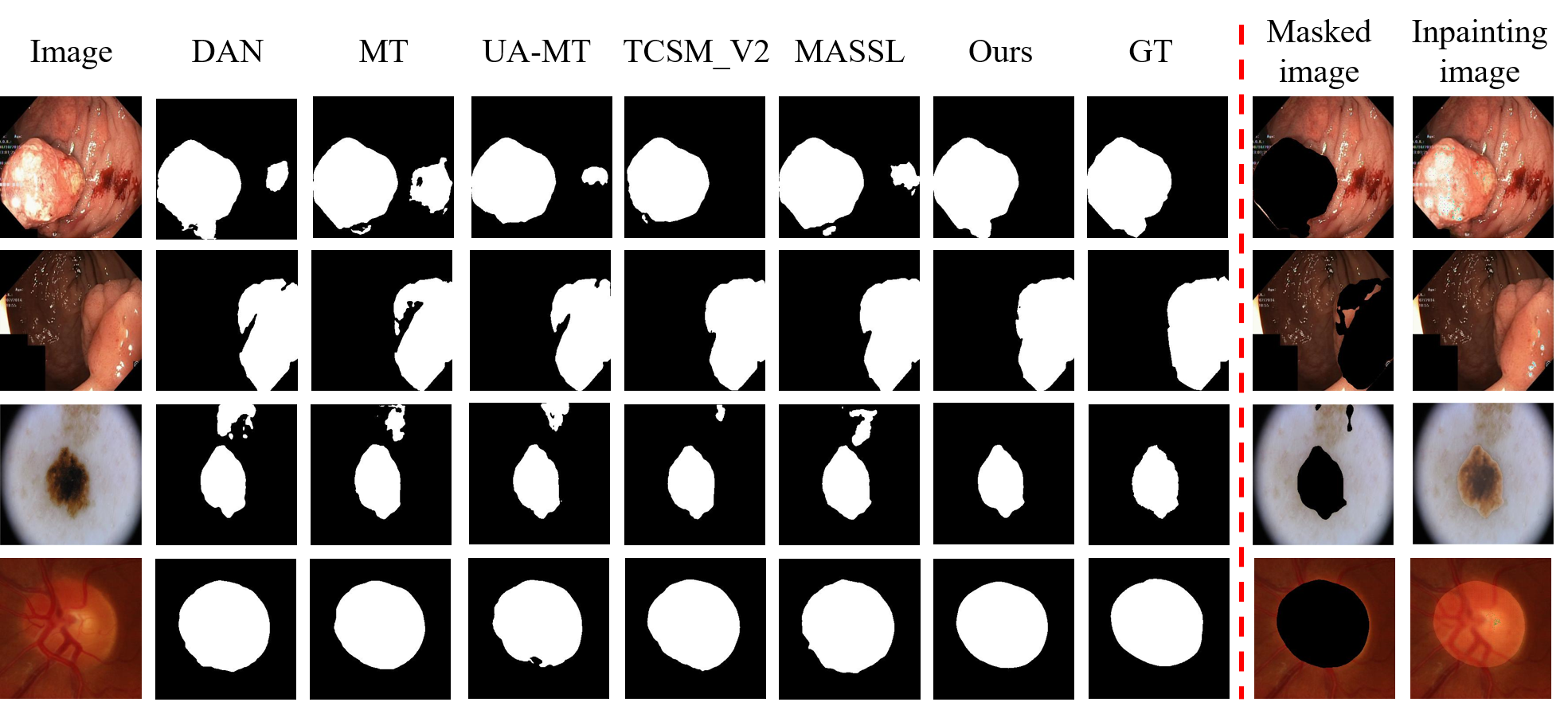}
\caption{Visual comparison of various lesion segmentation from state-of-the-art methods. Our proposed method consistently produces segmentation results closest to the ground truth. The inpainting result is shown in the rightmost column.}
\label{fig5}
\end{figure}

In our experiments, ResNet34~\cite{he2016deep} based UNet~\cite{ronneberger2015u} is utilized as our baseline, which is trained using all training set and our selected $20\%$ labeled data separately in a fully-supervised manner. Besides, we compare our method with other state-of-the-art approaches, including DAN~\cite{zhang2017deep}, MASSL~\cite{chen2019multi}, MT~\cite{tarvainen2017mean} and its variants (UA-MT~\cite{yu2019uncertainty}, TCSM\_V2~\cite{li2020transformation}). All comparison methods adopt ResNet34UNet as the backbone and use the same experimental settings for a fair comparison. On the \textbf{Kvasir-SEG dataset}, Table.~\ref{tab1} shows that our method obtains the outstanding performance compared with other semi-supervised methods, with Dice of $87.14\%$, which is $2.74\%$ improvement over the baseline only using the 120 labeled data, outperforming the second best method by $0.69\%$. On the \textbf{ISBI 2016 skin lesion dataset}, we obtain a $90.95\%$ Dice score, which is superior to other semi-supervised methods and very close to the score of $91.38\%$ achieved by the baseline using all training set images. On the \textbf{Rim-one r1 dataset}, we can conclude that our method achieves the best performance over five metrics, further demonstrating the effectiveness of our method. Note that detailed results on the latter two datasets are listed in the supplementary material due to the space limitation. Some visual segmentation results are shown in Fig.~\ref{fig5} (col.1-8).

\subsection{Ablation study}
\begin{figure}[htp]
\centering
\includegraphics[width=0.51\textwidth]{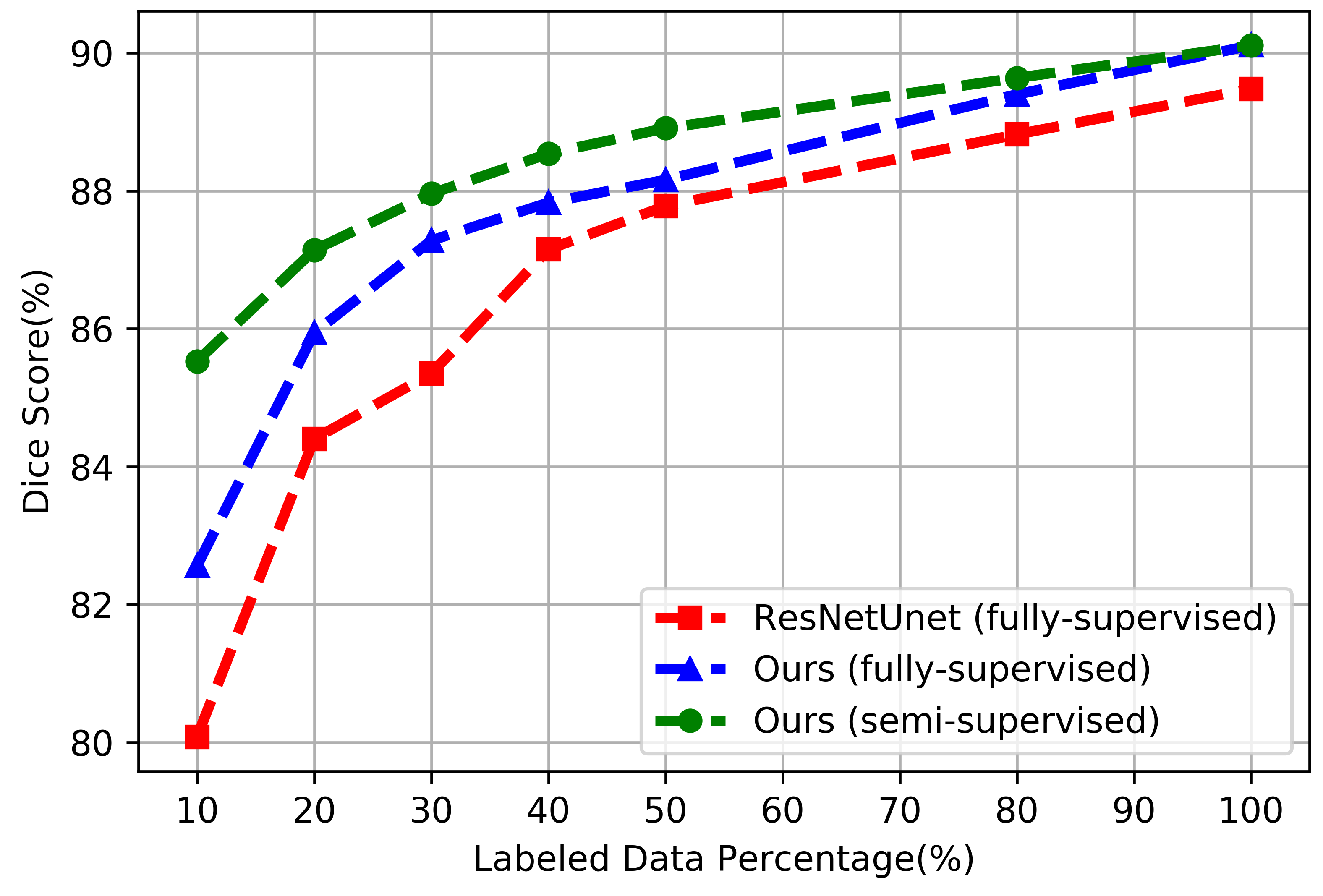}
\caption{The performance of our method with different ratio of labeled data on the Kvasir-SEG dataset.}
\label{fig4}
\end{figure}

\textbf{Effectiveness of our approach with different ratio of labeled data}
We draw the curves of Dice score under three settings in Fig.~\ref{fig4}. To verify that our proposed framework can mine residual lesion features and enhance the lesion representation by GFF modules in the second stage, we conduct experiments and draw the blue line. The blue line denotes that our method uses the same labeled data with the baseline (the red line) to perform the two-stage process, without utilizing any unlabeled data. Note that we only calculate the segmentation loss for the labeled data. The performance gains compared with the baseline show that our network mines useful lesion information in the second stage. The green line means that our method introduces the remaining as unlabeled data for the inpainting task, further enhancing the feature representation learning of the lesion regions and improving the segmentation performance, especially when only a small amount of labeled data is used. When using $100\%$ labeled data, the green line is equivalent to the blue line since no additional unlabeled data is utilized to do the inpainting task, thus maintaining the same results.\\
\textbf{Effectiveness of the GFF modules}
To verify the effectiveness of the GFF modules, we also design two variants, which merge features by directly addition and concatenation, denoting as Ours~(add) and Ours~(concat) respectively. In Table.~\ref{tab1}, we can observe performance degradation by both approaches compared with our method, proving that the GFF module plays a significant role in filtering redundant information and improving the model performance.  \\

\section{Conclusions}
In this paper, we believe that massive unlabeled data contains rich context and structural information, which is significant for lesion segmentation. Based on this, we introduce the self-supervised inpainting branch for unlabeled data, cooperating with the main segmentation task for labeled data, to further enhance the representation for lesion regions, thus refine the segmentation results. We also design the GFF module for better feature selection and aggregation from the two tasks. Experiments on various medical datasets have demonstrated the superior performance of our method.

\section*{Acknowledgement}This work is supported in part by the Key-Area Research and Development Program of Guangdong Province (No.~2020B0101350001), in part by the Guangdong Basic and Applied Basic Research Foundation (No.~2020B1515020048), in part by the National Natural Science Foundation of China (No.~61976250) and in part by the Guangzhou Science and technology project (No.~202102020633).

\bibliographystyle{splncs04}
\bibliography{paper867}

\end{document}